\documentclass[acmtog,screen,nonacm]{acmart}

\usepackage{multirow}
\usepackage{amsmath}
\usepackage{amssymb}
\usepackage{enumitem}
\usepackage{layouts}
\usepackage{hyphenat,balance,microtype}
\usepackage[fleqn,tbtags]{mathtools}
\usepackage[capitalise]{cleveref}
\usepackage[detect-weight]{siunitx}
\usepackage{caption,subcaption,textpos}
\usepackage{colortbl}

\usepackage{algorithm}
\usepackage{algorithmic}
\usepackage{verbatim}
\usepackage{listings}
\usepackage{wrapfig}

\graphicspath{{./images/}}

\definecolor{lightgray}{HTML}{F7F7F8}
\definecolor{tableheader}{HTML}{F3F4F6}
\definecolor{tablealt}{HTML}{FBFBFC}
\definecolor{tablegroup}{HTML}{F7F7F8}

\usepackage[most]{tcolorbox}
\usepackage{xcolor}

\newcommand{\tablestyle}[2]{\setlength{\tabcolsep}{#1}
                            \renewcommand{\arraystretch}{#2}
                            \centering
                            \footnotesize}

\definecolor{graycolor}{gray}{.9}

\setcopyright{none}
\usepackage{placeins}

\title{ViSculpt: Visual-Centric Agentic Geometry Editing}
\author{Bo Pang}
\authornote{Equal contribution.}
\affiliation{
  \institution{Peking University}
  \country{China}
}
\email{bo98@stu.pku.edu.cn}

\author{Jiaqi Pan}
\authornotemark[1]
\affiliation{
  \institution{Peking University}
  \country{China}
}
\email{panjiaqi@stu.pku.edu.cn}

\author{Xiaocheng Zhang}
\affiliation{
  \institution{Peking University}
  \country{China}
}
\email{zcadqewsxcsclc@163.com}

\author{Jiacheng Xu}
\affiliation{
  \institution{Peking University}
  \country{China}
}
\email{2401112075@stu.pku.edu.cn}

\author{Guoping Wang}
\authornote{Corresponding author. Peng-Shuai Wang is the project leader.}
\affiliation{
  \institution{Peking University}
  \country{China}
}
\email{wgp@pku.edu.cn}

\author{Peng-Shuai Wang}
\authornotemark[2]
\affiliation{
  \institution{Peking University}
  \country{China}
}
\email{wangps@hotmail.com}

\begin{abstract}
3D geometry editing is a critical yet labor-intensive part of the graphics pipeline, requiring artists to translate creative intent into precise operations in complex professional software.
Large language models (LLMs) have shown promise for script-based 3D creation, but script generation is less suited to perception-driven editing of arbitrary existing meshes, where execution must remain visually grounded and untouched regions should be preserved.
We present a \emph{visual-centric}, training-free multi-agent system that edits existing 3D meshes directly in Blender by emulating the iterative workflow of human artists.
Rather than generating scripts or regenerating geometry, our system operates through the Blender GUI: multimodal LLM agents observe the viewport, reason about the current mesh state, and execute localized edits through simulated user interactions.
Experiments on a curated benchmark provide initial evidence that this agentic approach can follow natural language instructions, perform representative localized mesh edits, and preserve the overall identity of the input asset.
Our results highlight a complementary regime for language-driven 3D editing: direct in-place modification of existing meshes within the native 3D editing workflow.
We view this work as an exploratory step toward visual-centric agentic geometry editing in professional graphics software.
\emph{We will release our code and benchmark to facilitate future research.}
\end{abstract}

\begin{teaserfigure}
  \centering
  \includegraphics[width=\linewidth]{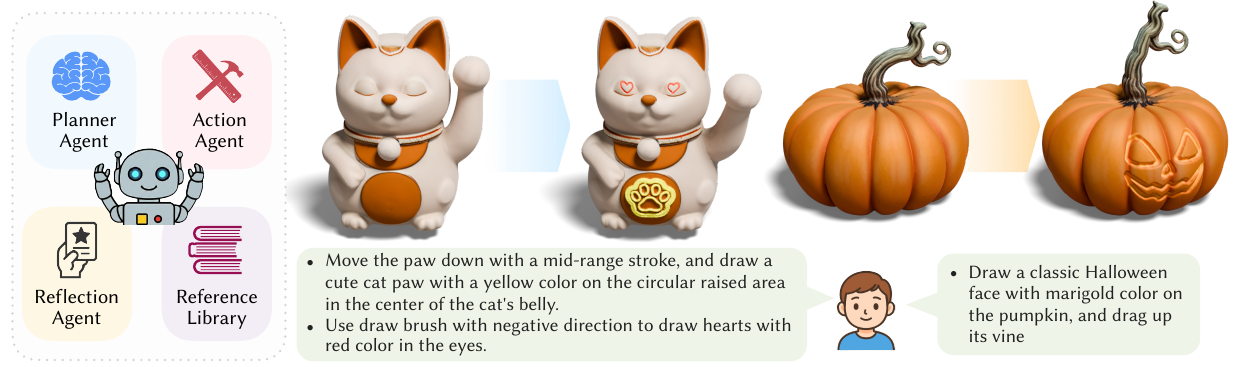}
  \caption{
  \textbf{Visual-Centric 3D Geometry Editing.}
  We present a training-free multi-agent system for editing existing 3D meshes from natural language.
  The system (left) plans editing steps, manipulates geometry directly through the Blender GUI, and visually evaluates intermediate results.
  The examples (right) show localized shape deformation and detailed surface sculpting on existing assets, while preserving the identity of the original mesh.
    }
  \label{fig:teaser}
\end{teaserfigure}

\begin{document}

\maketitle
\section{Introduction} \label{sec:intro}

3D geometry editing is central to computer graphics, enabling artists to reshape existing assets for design, animation, and fabrication.
It entails the precise manipulation of 3D shapes to achieve specific visual or functional goals.
For example, an artist may lengthen a creature's limbs for stylization or adjust a character's posture to convey a specific emotion.

High-quality geometry editing remains labor-intensive and demands domain expertise.
Professional artists typically manipulate 3D meshes in software such as Blender or Maya, translating high-level intent into long sequences of low-level operations while reasoning about topology, surface continuity, and visual appearance.
Recent diffusion-based generative methods~\cite{Zhang2024,Barda2024,gao20253d,Barda2025} can produce prompt-aligned 3D edits, but they often regenerate geometry rather than modify the given asset in place.
Consequently, they struggle to preserve instance-specific details and maintain high-fidelity correspondence to the input mesh.

Recent advances in large language models have enabled agents that interpret high-level intent, plan multi-step actions, and operate external tools~\cite{Gpt4,yao2022react,belle2025agents}.
These capabilities suggest a new path for geometry editing:
Can an LLM agent translate natural language instructions into executable operations by directly controlling professional 3D software?
If successful, such a paradigm would make sophisticated mesh editing accessible to users without specialized modeling expertise.

Recent work has shown that LLMs can drive 3D content creation by emitting API calls or Python programs for geometry construction~\cite{yuan20243d,lu2025ll3m,alrashedy2024generating,du2024blenderllm,raistrick2023infinite,raistrick2024infinigen}.
This program-generation paradigm is well suited to procedural assets and structured modeling workflows, and it can support edits when the underlying construction script remains available.
Yet it is less natural for direct, perceptually specified edits on arbitrary meshes: requests such as ``make the rabbit's ears more slender'' must be translated into precise local operations without continuous visual grounding in the modeling interface.
The mismatch is especially pronounced for existing 3D assets, which rarely provide a clean parametric representation; editing through the GUI offers a more direct and transparent handle on the visible geometry.

We introduce a \emph{visual-centric} paradigm for language-driven 3D editing, in which an LLM agent edits existing meshes by operating Blender much like a human artist: acting through the graphical user interface while continuously observing the evolving shape.
This formulation exploits the multimodal perception capabilities of modern LLMs~\cite{google2025gemini3pro} to interpret the current mesh state, assess the visual effect of each operation, and ground subsequent actions in immediate feedback.
The key technical obstacle is that professional modeling interfaces expose a vast, heterogeneous action space, making direct planning brittle even for strong models.
Our insight is that many sculpting operations can be factored into a compact set of primitive interaction patterns: \emph{Smear} for area coverage, \emph{Drag} for directed deformation, and \emph{Draw} for surface detail.
Each primitive is controlled by continuous parameters such as brush size, strength, anchor location, displacement, and stroke trajectory.
Our system translates high-level editing intent into sequences of these primitives and iteratively refines them from visual feedback.
By reducing GUI editing to a small yet expressive action vocabulary, the agent can perform localized modifications while preserving regions that should remain unchanged.
We do not claim that visual-centric interaction replaces the script-centric or generative approaches discussed above; rather, it occupies a complementary regime where the goal is to edit an existing mesh in place, preserve its identity, and make decisions from direct visual evidence.

Our system turns natural language into localized 3D mesh edits through a \emph{training-free} agentic pipeline.
It consists of a \emph{Planner Agent}, an \emph{Action Agent}, and a \emph{Reflection Agent}.
Given an instruction and an input mesh, the Planner Agent queries a sculpting knowledge base and decomposes the request into executable primitive actions.
The Action Agent performs these actions in Blender, selecting tools and adjusting parameters according to the current visual state.
The Reflection Agent acts as a visual critic, comparing intermediate results with the editing intent and deciding whether to refine the action or proceed.
Users may also provide feedback at any stage, allowing the plan to be revised during execution.
The system also stores successful editing trajectories, including tasks, actions, and visual outcomes, in an experience database, allowing the Planner Agent to retrieve and adapt effective strategies for future tasks.

We evaluate our approach on a curated benchmark of representative 3D editing tasks, with each task paired with before-and-after meshes created by professional artists.
User studies and evaluations across representative LLMs~\cite{anthropic2025claudesonnet45,seed2025seed18,Glm4.6v,openai2025gpt52,google2025gemini3pro,xai2025grok41,Qwen3-VL} show that our system can interpret natural language instructions and produce localized edits with competitive subjective ratings against human-edited references.
This work does not aim to replace expert artists or supplant script-centric and generative methods; instead, it establishes visual-centric agent interaction as a complementary path for direct, in-place editing of existing 3D assets.
We hope it opens a broader research direction at the intersection of agentic LLMs and graphics.
In summary, our main contributions are as follows:
\begin{itemize}[leftmargin=*, itemsep=2pt]
    \item[-] We introduce a visual-centric paradigm for language-driven editing of existing 3D meshes, formulating geometry editing as the feedback-driven interaction with professional graphics softwares.
    \item[-] We present a training-free multi-agent system built on compact primitive actions, enabling localized mesh edits through direct operation of Blender.
    \item[-] We provide initial empirical evidence, through qualitative results and blinded subjective evaluations, demonstrating promising edit quality while preserving the identity of the input mesh.
\end{itemize}

\section{Related Work} \label{sec:related}

\paragraph{3D Geometry Editing}
Geometry processing provides the algorithmic foundation for manipulating and analyzing 3D data~\cite{Botsch2010}, including mesh simplification~\cite{Garland1997,Hoppe1999}, remeshing~\cite{Alliez2003,Yan2009}, surface smoothing~\cite{Desbrun1999}, shape deformation~\cite{Sorkine2007,Sorkine2004}, and correspondence~\cite{Ovsjanikov2012}.
Within this area, \emph{geometry editing} refers to tasks that modify a shape to satisfy user-specified requirements.
Classical editing methods are typically driven by geometric signals such as curvature, topology, or error metrics, and thus provide limited support for high-level semantic intent.
Recent generative methods~\cite{Zhang2024,Barda2024,gao20253d,Barda2025,Li2024a} learn to produce edited shapes conditioned on an input shape and an edit specification.
These methods are powerful for generation and representation-level transformation, but our setting instead requires direct, localized edits to an existing mesh while preserving unintended regions.
We therefore view generative editing as a complementary direction.
Our work targets the remaining gap between semantic editing intent and expert GUI operation by enabling visual-centric geometry editing in professional 3D software directly from natural language.

\paragraph{3D Agentic System}
Advances in LLMs~\cite{Gpt4,seed2025seed18,Glm4.6v,openai2025gpt52,google2025gemini3pro,xai2025grok41,Qwen3-VL} have enabled agentic systems that interleave planning, perception, and tool use.
In 3D content creation, prior agents commonly translate natural-language instructions into executable programs, including Blender Python scripts, procedural programs, and parametric modeling code~\cite{yuan20243d,lu2025ll3m,alrashedy2024generating,du2024blenderllm,raistrick2023infinite,raistrick2024infinigen,yamada2025l3go,Ahuja2025blendermcp,hu2024scenecraft,Lv2024,Sun2025,huang2024}.
Related work further improves controllability through domain-specific languages that expose semantic 3D concepts as compositional primitives~\cite{Sharma2018,kania2020ucsg,du2018inversecsg,wu2021deepcad,chen2025img2cad,jones2020shapeassembly}, or by coupling perception and planning with low-level geometric operations~\cite{liu2025,fu2024scene,ling2025}.
Most of these systems follow a script-generation paradigm, targeting asset creation or editing through an intermediate program representation.
Our work is complementary: we study direct GUI-based editing in professional 3D software, focusing on existing meshes, visual feedback, localized operations, and preservation of untouched regions.
\looseness=-1

\paragraph{Multimodal Large Models}
Recent multimodal foundation models can integrate visual and linguistic signals for complex tasks~\cite{Radford2021,li2022n,lu2025ll3m,Krishna2023}, providing a general interface for grounding language in visual observations and 3D scenes.
In 3D generation, vision-language models (VLMs) often serve as semantic priors for optimizing implicit representations~\cite{Zhang2024,Chen2024s}; for example, CLIP-Mesh~\cite{mohammad2022} uses VLM-based discrimination over rendered views to guide text-consistent geometry synthesis.
Segmentation foundation models such as Segment Anything (SAM)~\cite{kirillov2023,ravi2024,wei2024n,shen2023} are likewise used in 3D reconstruction and to transfer strong 2D priors to 3D settings~\cite{cen2023,yang2023}.
Most prior work, however, applies these models to generation or reconstruction rather than interactive editing.
We instead use VLMs and SAM-like segmentation models to ground visual-centric geometry editing inside professional 3D software, aligning localized operations with user intent.

\section{Method} \label{sec:method}

\begin{figure}[t]
  \centering
  \includegraphics[width=0.99\linewidth]{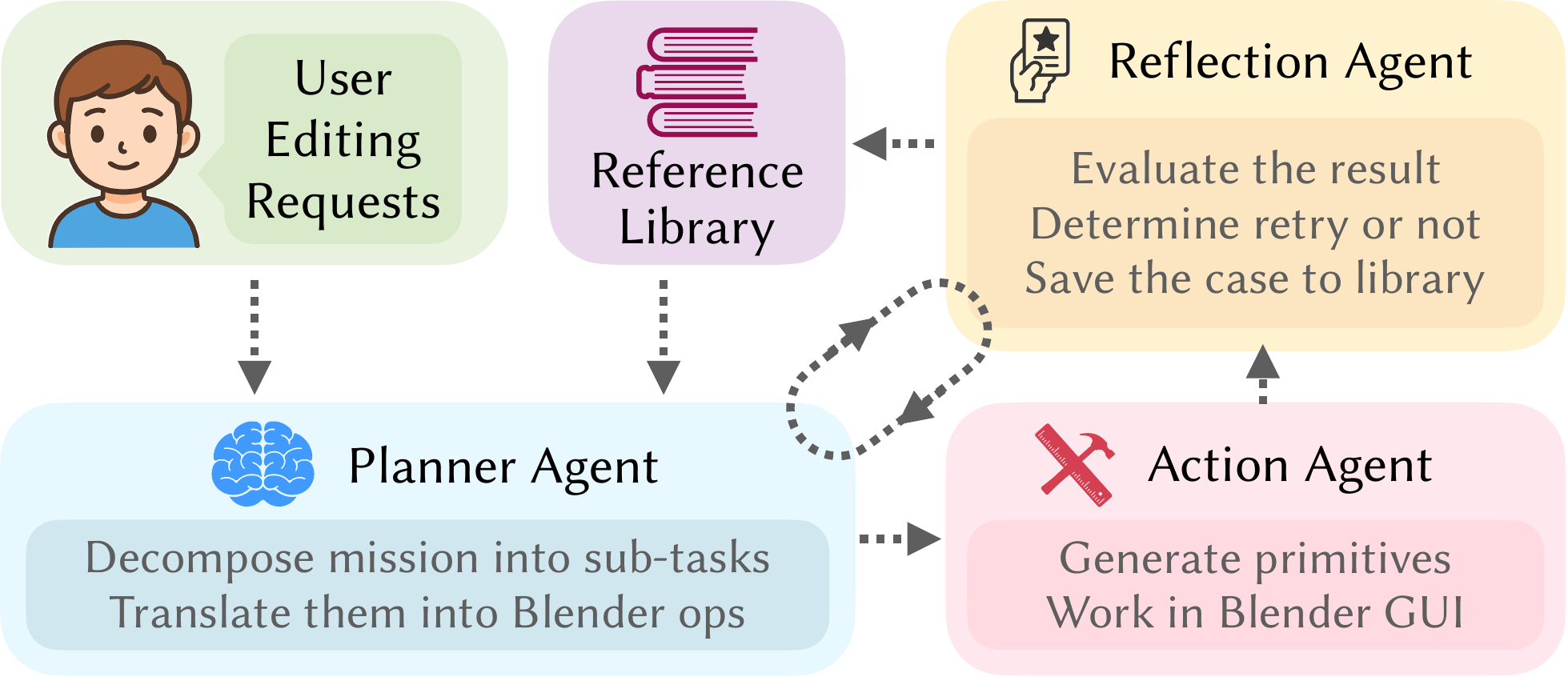}
  \caption{
  Overview of the proposed agent system. The Planner Agent decomposes a natural language instruction into executable steps, the Action Agent performs these steps in Blender, and the Reflection Agent evaluates the result and provides feedback. The system supports both local refinement loops and long-term improvement through reference libraries.
  }
  \Description{}
  \label{fig:overview}
\end{figure}

Given an input 3D mesh and a natural language description of the desired changes, our goal is to automate 3D geometry editing by producing a modified mesh that aligns with the user's editing intent.
Our system is training-free and fully automatic, while still allowing users to provide feedback at any stage.
To this end, we introduce a multi-agent system built on primitive editing actions, with three components: a Planner Agent, an Action Agent, and a Reflection Agent.
An overview of the system is shown in \cref{fig:overview}.
We first describe the primitive mouse actions that form the basis of our editing operations in \cref{sec:three_primitive_actions}, then detail how the agents compose a coherent editing pipeline in \cref{sec:model_pipeline}.

\subsection{Primitive Actions}\label{sec:three_primitive_actions}

\begin{figure*}[t]
  \centering
  \includegraphics[width=0.99\linewidth]{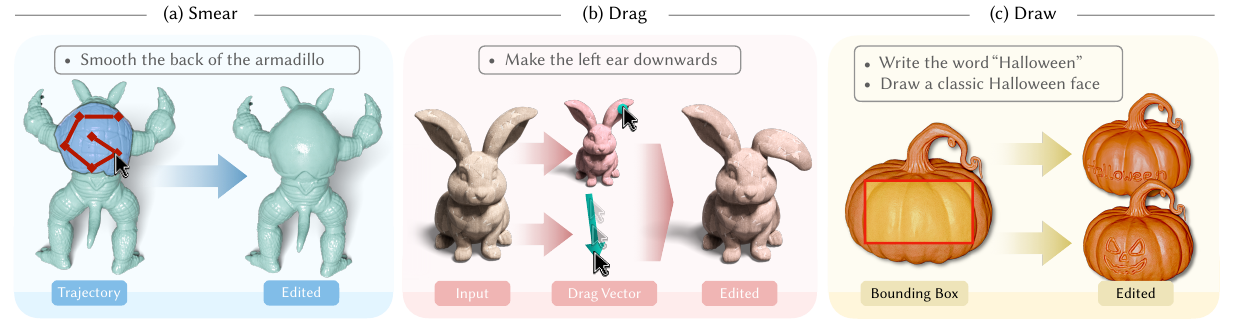}
  \caption{Three primitive mouse trajectories. (a) Smear, generated with segmentation and morphological operations; (b) Drag, defined by an anchor point from QuadLoc and a VLM-predicted displacement vector; and (c) Draw, created through heuristic stroke extraction for text and contour extraction for arbitrary shapes.
  }
  \label{fig:three_primitives}
  \Description{}
\end{figure*}

In Blender sculpting, brush editing operations often depend on precise mouse trajectories.
Directly generating such trajectories with LLM agents is difficult because the action space is continuous and high-dimensional.
We observe that many brush strokes can be represented by three primitive mouse trajectories: \emph{Smear}, \emph{Drag}, and \emph{Draw}.
This abstraction constrains generation to a compact action vocabulary while remaining expressive enough for the representative brush operations considered in this work.
\Cref{fig:three_primitives} summarizes the three primitives.
We next describe how each trajectory is generated.

\paragraph{Smear}
The \emph{Smear} trajectory implements area filling for brushes that require near-uniform coverage over a surface region, such as smoothing or texture painting.
We first use a VLM to coarsely localize the target region in the current Blender viewport and produce a bounding box.
This box prompts the Segment Anything Model (SAM)~\cite{kirillov2023} to generate a binary mask, from which we extract the exterior boundary using OpenCV.
For \emph{Smear}, the LLM agent only specifies the brush radius; the trajectory is then constructed deterministically from the mask.
Given the radius, we iteratively erode the mask to obtain nested contours, sample points along them, and connect the samples into a continuous outside-in filling stroke that maintains consistent coverage.

\begin{figure}[t]
  \centering
  \includegraphics[width=0.99\linewidth]{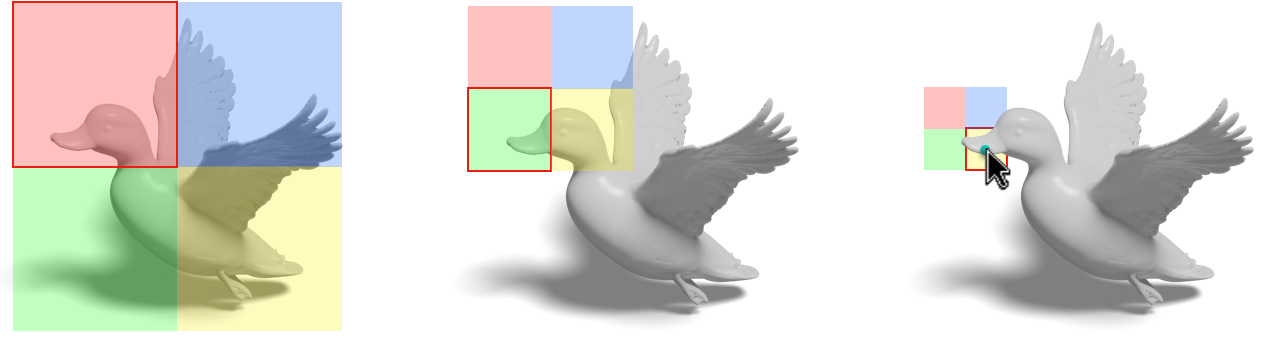}
  \vspace{-2mm}
  \caption{QuadLoc procedure.
    To mitigate VLM spatial imprecision, QuadLoc recursively queries the model for the quadrant containing the target (red outline), progressively narrowing the search space from left to right.
  }
  \label{fig:quadloc}
\end{figure}

\paragraph{Drag}
The \emph{Drag} trajectory supports local deformation by pulling or pushing a selected part of the model along a specified displacement.
It is central to manipulations such as limb elongation, pose adjustment, and protrusion creation.
We parameterize \emph{Drag} by an anchor point and a displacement vector.
The anchor requires accurate localization of a mesh feature,
but directly querying even advanced VLMs~\cite{google2025gemini3pro} for image-space coordinates is unreliable and often yields inconsistent or hallucinated positions.
We therefore introduce \emph{QuadLoc}, a coarse-to-fine quadrant localization scheme that casts localization as a sequence of visual multiple-choice decisions.
Given the current viewport image, we overlay a $2 \times 2$ grid with four uniquely colored quadrants (Red, Blue, Green, Yellow) and ask the VLM to select the quadrant containing the target anchor, instead of predicting exact coordinates, as shown in the left part of \cref{fig:quadloc}.
We crop to the selected quadrant and repeat the query recursively until the quadrant size falls below a preset threshold.
The center of the final quadrant is used as the anchor point.
The displacement vector encodes direction and magnitude and is inferred by prompting a VLM to interpret the requested deformation; in practice, we find this prediction reliable.

\paragraph{Draw}
The \emph{Draw} trajectory handles high-frequency surface details, such as inscribed text or intricate patterns.
Unlike \emph{Smear} and \emph{Drag}, \emph{Draw} requires stroke-level precision.
We use different strategies for \emph{text} and \emph{arbitrary shapes}.
For text, we synthesize handwriting-like paths by rasterizing the target string into a binary mask using the handwriting-style font ``Patrick Hand''.
We compute the medial axis of the glyph mask with \texttt{scikit-image}, convert the skeleton into a topological graph, and traverse graph edges in an order that approximates handwriting dynamics.
For arbitrary shapes, such as symbols, emojis, or logos, we use a 2D image generation model to synthesize a high-contrast binary reference from the user's description.
We then extract salient contours from the generated image and map them to viewport coordinates as drawing strokes.

\subsection{Agent System for Geometry Editing}\label{sec:model_pipeline}

We next describe the three specialized agents in our editing system.
For clarity, we use a running example in which the user asks to ``pull the man's arms inward naturally and write \texttt{SIGGRAPH} on his chest'' for a 3D human mesh, as shown in \cref{fig:example_pipeline}.

\subsubsection{Planner Agent}\label{sec:planner_agent}

The Planner Agent translates high-level, potentially ambiguous user intent into an ordered list of executable commands.
It has two components: a \emph{Decomposer} and a \emph{Translator}.

\begin{figure*}[t]
  \centering
  \includegraphics[width=0.98\linewidth]{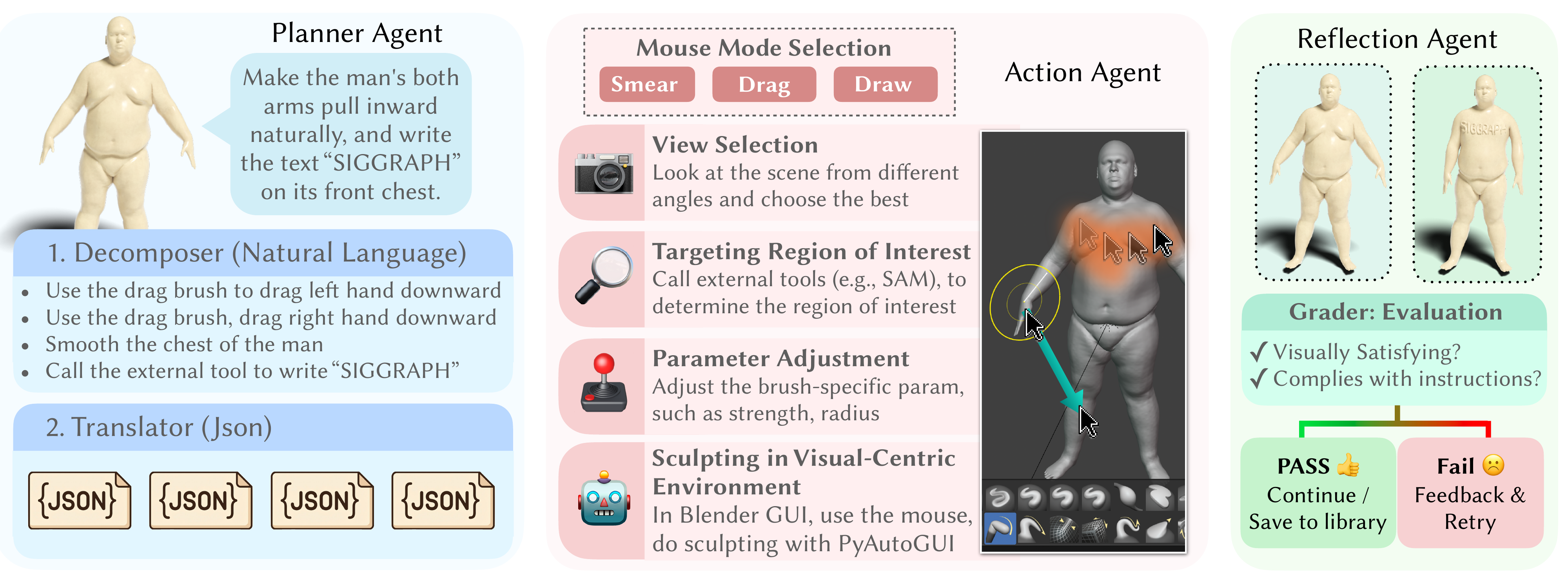}
  \caption{
  Example of our agentic editing pipeline. The Planner Agent decomposes a natural language request into sub-tasks and structured JSON commands. The Action Agent executes each command in Blender by selecting a primitive action, choosing a view, localizing the target region, and simulating GUI interactions. The Reflection Agent evaluates the result against the instruction, triggering refinement when needed and archiving successful trajectories for future reuse.}
  \label{fig:example_pipeline}
  \Description{}
\end{figure*}

\paragraph{Decomposer}
The Decomposer performs semantic parsing and task planning from the current mesh state, the user instruction, and retrieved reference information.
Let $\mathcal{G}$ denote the current mesh, $\mathcal{D}_0$ the initial user instruction, and $\mathcal{R}$ the information retrieved from the Reference Library (\cref{sec:reference_library}).
The Decomposer $\mathbf{P_d}$ produces a structured plan $\mathcal{D}_1 = \mathbf{P_d}(\mathcal{G}, \mathcal{D}_0, \mathcal{R})$.
The plan $\mathcal{D}_1$ is an ordered list of sub-tasks that disambiguate the intended operations and their dependencies.
In our running example, the Decomposer splits the request into distinct actions shown on the left of \cref{fig:example_pipeline}.

\paragraph{Translator}
While the Decomposer determines \emph{what} to do, the Translator determines \emph{how} to execute it.
The Translator $\mathbf{P_t}$ converts $\mathcal{D}_1$ into machine-readable specifications for the Action Agent.
It uses the Blender API documentation from the Reference Library $\mathcal{R}$ and optional iterative feedback $\mathcal{F}$ from the Reflection Agent (\cref{sec:reflection_agent}) to generate JSON commands:
$ \mathbf{J} = \mathbf{P_t}(\mathcal{D}_1, \mathcal{F}, \mathcal{R}) $.
The resulting specification $\mathbf{J}$ contains the attributes needed to execute the operation and can be refined at execution time using visual observations.
In our implementation, most refinements occur at the execution level, such as adjusting localization or parameters, or retrying the same sub-task, rather than re-planning the entire task.

\subsubsection{Action Agent}\label{sec:action_agent}

The Action Agent executes the Planner's JSON commands $\mathbf{J}$ by interacting with the Blender GUI.
Rather than relying on script-based API calls, it is \emph{visual-centric}: it perceives Blender's viewport and synthesizes mouse events in a human-like interaction loop.
The execution proceeds in three stages: \emph{View Selection}, \emph{Target Segmentation}, and \emph{GUI Control}.
Refer to the middle of \cref{fig:example_pipeline} for an illustration of these stages.

\paragraph{View Selection}
Before editing, artists adjust the viewpoint to expose the target area.
Similarly, the agent selects a view that maximizes target visibility and reduces occlusion.
It renders six canonical views ($+x, -x, +y, -y, +z, -z$) and prompts the VLM to choose the view with the clearest visibility of the region referenced by the Planner.
Although free camera control is more expressive, current foundation models often lack robust 3D spatial reasoning; unconstrained camera manipulation can cause the agent to lose track of the object while consuming excessive context.
For most object-centric assets without severe self-occlusion, selecting among six canonical views provides a reliable and efficient proxy for full viewport navigation. \looseness=-1

\paragraph{Target Segmentation}
The agent localizes the region of interest with a coarse-to-fine pipeline.
It captures the selected viewport, queries the VLM for a bounding box around the instructed target, and uses this box to prompt a Grounded Segment Anything Model~\cite{ravi2024} to produce a binary mask.
The mask constrains subsequent operations to the intended region, reducing unintended edits to unrelated geometry.

\paragraph{GUI Control}
Given the selected view and mask, the agent executes the operation through screen-space manipulation of the Blender GUI.
Using \texttt{PyAutoGUI}, it configures brush parameters from the JSON command, maps primitive trajectories (Smear, Drag, Draw) from image space to viewport coordinates, and simulates mouse button states (press/hold/release).
As the interaction uses Blender's native stroke processing, it inherits built-in stabilization and pressure simulation, yielding edits consistent with  human workflows. \looseness=-1

\subsubsection{Reflection Agent}\label{sec:reflection_agent}

Multi-step editing without verification can accumulate errors.
We therefore introduce a \emph{Reflection Agent} that evaluates each executed sub-task and decides whether to accept the result or request refinement, as shown in the right part of \cref{fig:example_pipeline}.

To assess edit quality, we employ a VLM-based Visual Grader.
The Grader takes a pre-edit screenshot $\mathcal{I}_{pre}$, a post-edit screenshot $\mathcal{I}_{post}$, and the Planner-generated sub-goal $\mathcal{D}_{sub}$ as input.
It compares the visual change from $\mathcal{I}_{pre}$ to $\mathcal{I}_{post}$ with the semantic intent of $\mathcal{D}_{sub}$ using a visual question-answering formulation.
The Grader outputs a scalar score $S \in [0, 15]$ and a textual critique.
Given a threshold $\tau$, we accept the sub-task if $S \geq \tau$ and advance to the next sub-task; accepted instances are archived in the Previous Success Cases library to inform future decisions.
If $S < \tau$, the agent enters a \emph{Refinement Loop}: the critique is provided as feedback context $\mathcal{F}$ to the Planner and Action Agent, which revise parameters (e.g., increasing brush strength) and re-execute the sub-task.
We emphasize that this grader is used as a practical visual critic for iterative refinement, rather than as a definitive objective measure of geometric correctness.

\subsubsection{Reference Library} \label{sec:reference_library}

\begin{figure}[htbp]
  \centering
  \includegraphics[width=0.99\linewidth]{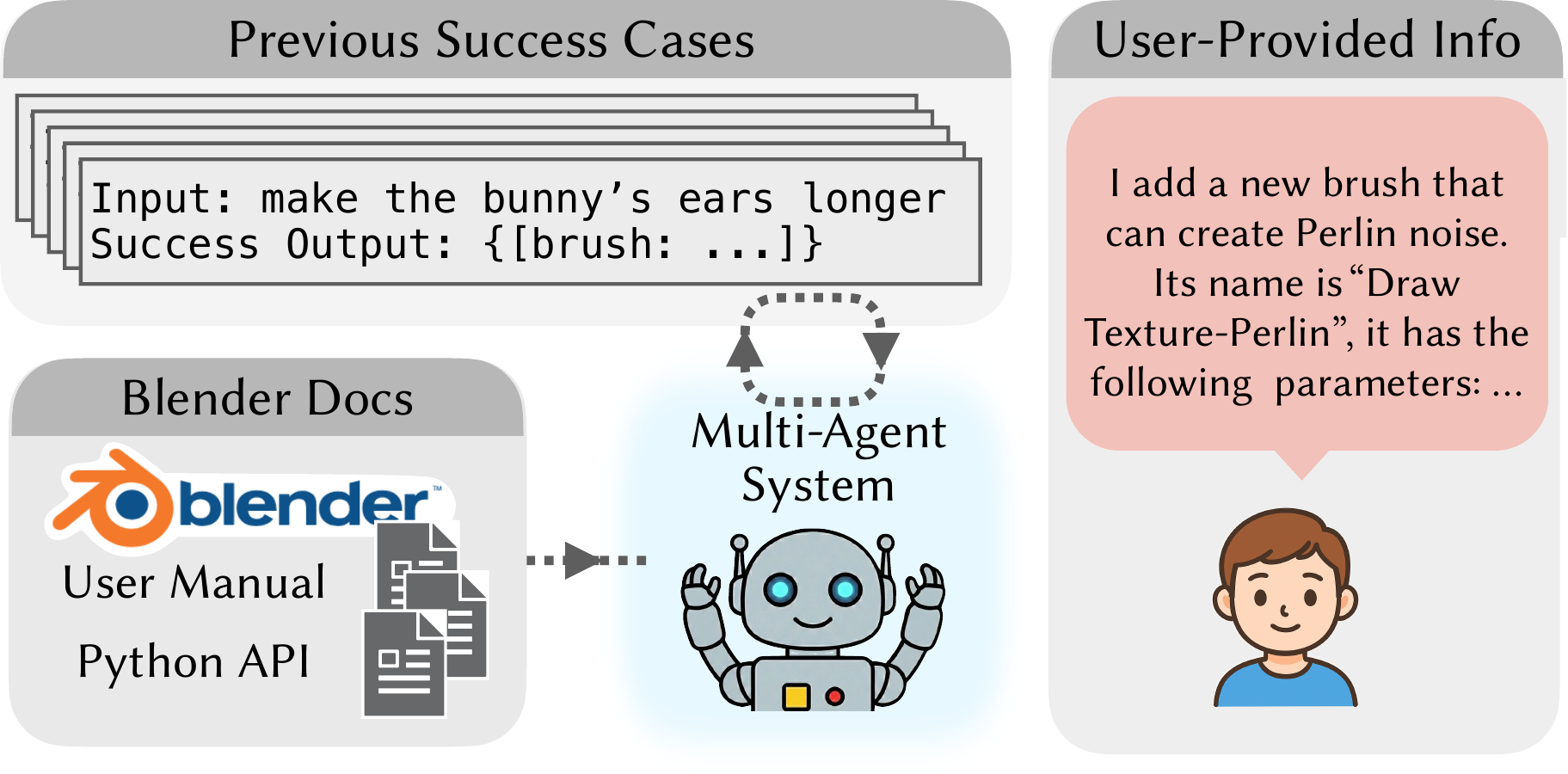}
  \caption{The Reference Library augments the
multi-agent system via Retrieval-Augmented Generation (RAG). It integrates knowledge from three sources: Previous Success Cases (accumulated runtime
experience), Blender Docs (official software manuals), and User-Provided Info (custom tool definitions).
}
  \label{fig:rag}
  \Description{}
\end{figure}

To equip the agents with external knowledge, we maintain a Reference Library $\mathcal{R}$ queried via retrieval-augmented generation (RAG), as shown in \cref{fig:rag}, which comprises three sub-libraries:
\begin{itemize}[leftmargin=*, itemsep=2pt]
    \item[-] \emph{Blender Knowledge Library} stores official documentation for  sculpting tools and Python APIs of Blender. The Translator queries this library to ensure that generated   compatible JSON commands.
    \item[-] \emph{User-Provided References} store user-specific assets (e.g., custom brushes), user preferences, and style guidelines.
    \item[-] \emph{Previous Success Cases} store successful past operations. For a new task, the Planner retrieves the top five most similar cases to guide planning and translation.
\end{itemize}

The Blender Knowledge Library and User-Provided References are populated before deployment, whereas Previous Success Cases are updated online during execution.
When the Grader assigns a high score to a completed task, we archive the corresponding JSON command and execution context, enabling the agent to reuse effective strategies in subsequent tasks.
Thus, the agent improves over time by accumulating high-quality exemplars.

\section{Results and Comparisons}
\label{sec:result}

\paragraph{Environment Setup}
We use Google Gemini 3 Flash as the default LLM/VLM backbone for all three agents.
For the Action Agent's segmentation module, we use Gemini 3 Pro for stronger visual grounding and Grounded SAM2~\cite{ravi2024sam2} for mask generation.
For free-form Draw actions, we use Z-Image~\cite{cai2025z} to generate reference images.
We access Gemini models through the official API.
To demonstrate extensibility, we also include custom procedural brushes, including star-pattern and Perlin-noise brushes.
All experiments are conducted on Windows 11 with Blender 4.5 LTS.
In practice, an edit takes about 2 minutes, while multi-step revisions take about 8 minutes, comparable to~\cite{lu2025ll3m}.
Notably, more than 85\% of the total runtime is attributed to the latency of LLM inference; faster model serving would directly improve interactivity.
\looseness=-1

\paragraph{Benchmark and User Study}
Direct 3D geometry editing remains a challenging task, with few automatic systems supporting localized, instruction-driven changes to existing meshes.
We therefore evaluate our agent-based system against human artists using a blinded user study.
We curate 20 geometry-editing tasks and produce two results for each task: one generated by our system and one manually created in Blender by human artists, yielding 40 edited meshes in total.
Participants viewed randomized pre-/post-edit mesh pairs together with the corresponding instruction, and rated each result from 0 to 10 for instruction adherence, visual quality, and geometric plausibility.
The study included 48 participants: 39 non-experts and 9 experts, with experts defined as participants with prior publication experience in computer graphics venues.
Since visual-centric editing of existing meshes is a relatively new setting, large-scale standardized comparisons against multiple automatic baselines are currently difficult to establish fairly across the same task scope, software environment, and evaluation protocol.
We therefore use human-created edits as the primary reference, providing a direct and stable comparison for the in-place editing tasks studied here.

\begin{figure*}[t]
  \centering
  \includegraphics[width=\textwidth]{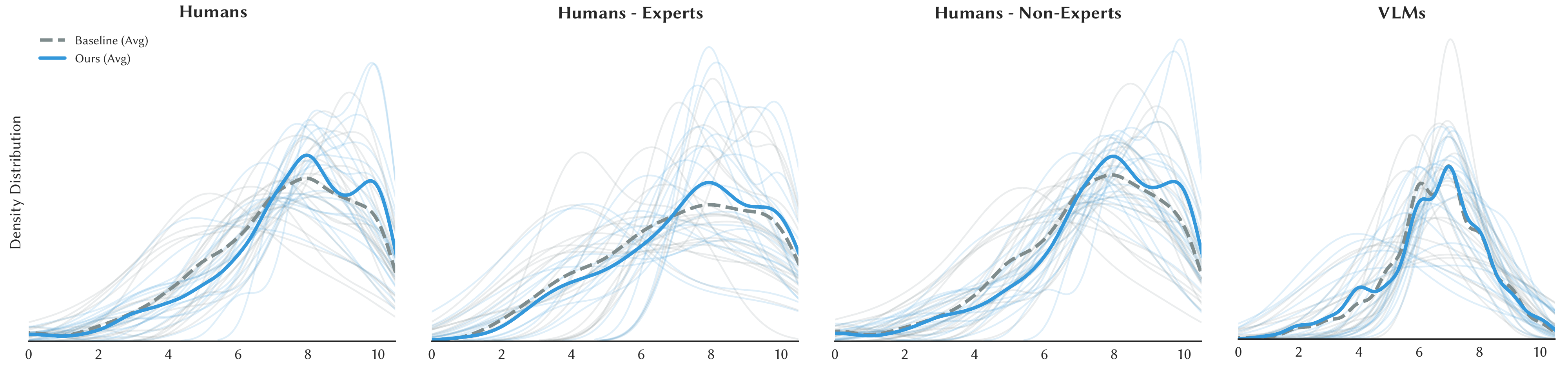}
  \caption{Visualization of score distributions across all 20 test examples. The plots display the Kernel Density Estimation (KDE) of user scores (0-10 scale) for our method (blue) and the baseline (gray).
  Thin semi-transparent lines represent the score distribution for each individual example, illustrating the performance consistency across diverse inputs.
  Thick lines indicate the aggregated average distribution of each group.}
  \label{fig:user_study_figure}
\end{figure*}

The results of the user study are summarized in \cref{tab:user_study_results} and \cref{fig:user_study_figure}.
Our method achieves an average score of 7.53, comparable to the 7.20 average score of human-created edits.
This result indicates that our visual-centric agent pipeline can produce perceptually plausible geometry edits across the evaluated task set.
We further conduct an auxiliary VLM-based evaluation using the same questionnaire, repeating each evaluation five times to reduce sampling variance.

\begin{figure*}[p]
  \centering
  \includegraphics[width=\textwidth,height=0.94\textheight,keepaspectratio]{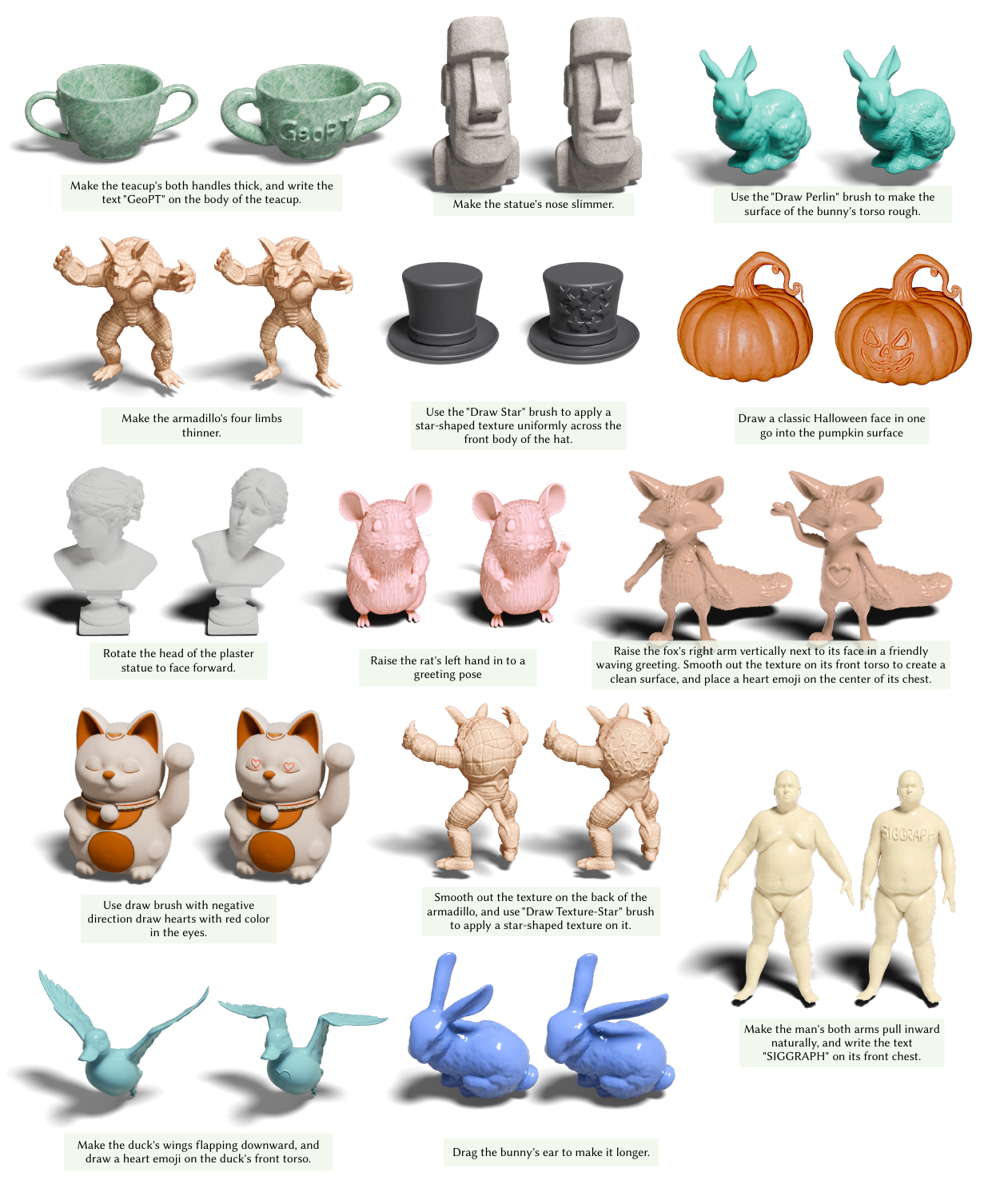}
  \caption{Results of our method on various 3D mesh editing tasks.
  For each example, we show the input mesh on the left, the natural language description of the desired changes in the bottom, and the edited mesh generated by our method on the right.}
  \label{fig:experiment_samples}
\end{figure*}

\begin{figure*}[t]
  \centering
  \includegraphics[width=\textwidth,height=0.72\textheight,keepaspectratio]{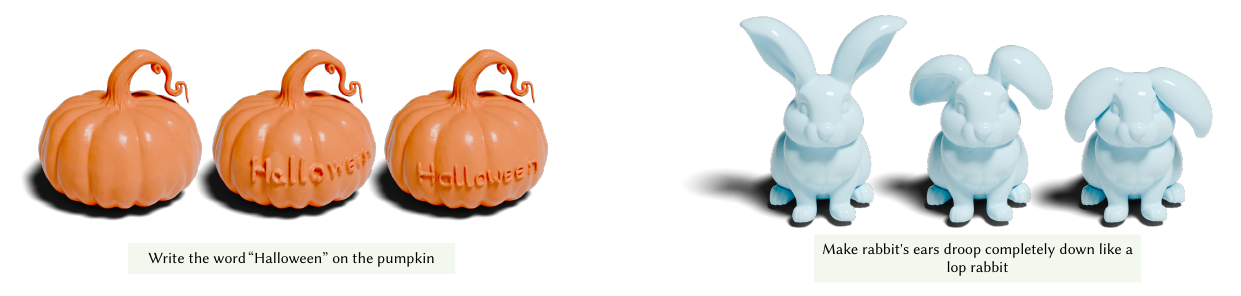}
  \caption{
    The comparison between our method and human editing.
    In each example, we show the input mesh on the left, the edited mesh generated by our method on the right, and the edited mesh created by human artists in the middle.
  }
  \label{fig:agent_vs_human}
\end{figure*}

We show some qualitative results of our method in \cref{fig:experiment_samples}.
In each example, we show the input mesh, the natural language description of the desired changes, and the edited mesh generated by our method.
Representative comparisons between our agent and human artists are visualized in \cref{fig:agent_vs_human}.
Together, these examples illustrate the localized editing operations targeted by our system and complement the quantitative user-study results.

\paragraph{Comparison with Human Artists}
Compared with manual editing by human artists, our method substantially reduces the interaction burden while achieving competitive perceptual quality.
For instance, painting a smiley face on a mesh may require an artist to prepare a texture in an external image editor and configure Blender's stencil-mapping workflow.
With our system, the user can express the same intent directly in natural language, e.g., ``paint a smiley face on the rabbit''.
For edits that imply symmetry, such as ``droop both ears,'' the agent also infers the corresponding brush configuration instead of requiring manual tool setup.
Our current implementation is not intended to replace expert production workflows: it may be slower than skilled manual editing in some cases, primarily due to model-inference latency.
Rather, it demonstrates a path toward accessible geometry editing in which non-expert users can perform localized mesh edits through high-level visual intent.

\begin{table}[t]
  \caption{User study statistics comparing our results with human-created edits. Scores are reported as mean and standard deviation (SD) across all participants and tasks, where each score is given on a 0--10 scale.}
  \label{tab:user_study_results}
  \tablestyle{5pt}{1.2}
  \begin{tabular}{lccc}
    \toprule
    & & \multicolumn{2}{c}{\textbf{Mean Score} $\pm$ SD} \\
    \cmidrule(lr){3-4}
    \textbf{Group} & \textbf{N} & \textbf{Human Reference} & \textbf{Ours} \\
    \midrule
    \rowcolor{tablegroup}
    \textbf{All Users} & \textbf{48} & $\textbf{7.20} \pm \textbf{2.19}$ & $\textbf{7.53} \pm \textbf{2.13}$ \\

    \rowcolor{white}
    \ \ Experts & 9 & $7.12 \pm 2.23$ & $7.36 \pm 2.17$ \\
    \rowcolor{tablealt}
    \ \ Non-Experts & 39 & $7.22 \pm 2.19$ & $7.57 \pm 2.12$ \\

    \rowcolor{tablegroup}
    \textbf{All VLMs} & \textbf{55} & $\textbf{6.54} \pm \textbf{1.01}$ & $\textbf{6.51} \pm \textbf{0.87}$ \\

    \rowcolor{white}
    \ \ Claude 4.5 Opus, Sonnet~\shortcite{anthropic2025claudesonnet45} & 10 & $5.51 \pm 0.54$ & $5.68 \pm 0.18$ \\
    \rowcolor{tablealt}
    \ \ Doubao Seed 1.8~\shortcite{seed2025seed18} & 5 & $7.84 \pm 0.16$ & $7.17 \pm 0.08$ \\
    \rowcolor{white}
    \ \ GLM 4.6V~\shortcite{Glm4.6v} & 5 & $6.77 \pm 0.17$ & $6.56 \pm 0.20$ \\
    \rowcolor{tablealt}
    \ \ GPT 5.2, 5.2 Pro~\shortcite{openai2025gpt52} & 10 & $5.93 \pm 0.09$ & $5.90 \pm 0.06$ \\
    \rowcolor{white}
    \ \ Gemini 3 Pro, Flash~\shortcite{google2025gemini3pro} & 10 & $7.28 \pm 0.91$ & $7.26 \pm 0.83$ \\
    \rowcolor{tablealt}
    \ \ Grok 4~\shortcite{xai2025grok4}, 4.1-fast~\shortcite{xai2025grok41} & 10 & $7.01 \pm 1.20$ & $7.17 \pm 0.98$ \\
    \rowcolor{white}
    \ \ Qwen3 VL Plus~\shortcite{Qwen3-VL} & 5 & $5.85 \pm 0.31$ & $5.81 \pm 0.30$ \\
    \bottomrule
  \end{tabular}

\end{table}

\begin{table}[t]
  \caption{Design-space comparison of representative paradigms for language-driven 3D editing.
  Here $\checkmark$ denotes a natural strength, $\triangle$ denotes partial or scenario-dependent support, and $\times$ denotes that the property is typically not a primary strength.
   }
  \label{tab:design_space_comparison}
  \tablestyle{4pt}{1.3}
  \begin{tabular}{p{0.48\linewidth}ccc}
    \toprule
    \textbf{Property} & \textbf{Script-centric} & \textbf{Generative} & \textbf{Ours} \\
    \midrule
    \rowcolor{white}
    Direct in-place editing on existing mesh & $\triangle$ & $\times$ & $\checkmark$ \\
    \rowcolor{tablealt}
    Preserve untouched regions & $\triangle$ & $\times$ & $\checkmark$ \\
    \rowcolor{white}
    Immediate visual feedback & $\times$ & $\times$ & $\checkmark$ \\
    \rowcolor{tablealt}
    Works without construction history & $\triangle$ & $\checkmark$ & $\checkmark$ \\
    \rowcolor{white}
    Natural fit for localized perceptual edits & $\triangle$ & $\times$ & $\checkmark$ \\
    \rowcolor{tablealt}
    Native interaction inside 3D software & $\triangle$ & $\times$ & $\checkmark$ \\
    \rowcolor{white}
    Open-ended geometry synthesis & $\triangle$ & $\checkmark$ & $\triangle$ \\
    \bottomrule
  \end{tabular}
\end{table}

\section{Discussion and Ablation Studies}

We compare our visual-centric editing paradigm with script-centric and generative alternatives, clarifying the operating regimes and trade-offs of each family, which are summarized in \cref{tab:design_space_comparison}.
We then ablate our primitive abstraction to evaluate its role in stabilizing execution and grounding semantic instructions into precise GUI actions.
\looseness=-1

\paragraph{Discussion: Script-Centric Approaches}
Script-centric systems~\cite{lu2025ll3m,du2024blenderllm} provide a natural comparison with our approach because they also connect language models to professional 3D software.
These methods are well suited to tasks with explicit procedural structure, construction history, or programmatic decomposition.
Our setting targets a complementary regime: localized edits to \emph{existing} meshes, where human instructions are often perceptual and preserving untouched regions is as important as modifying the target region.
Rather than synthesizing a program or regenerating geometry from an intermediate representation, our agent acts directly on the mesh state observed in the GUI, using visual feedback to perform in-place edits while keeping the surrounding asset stable.
\looseness=-1

\begin{figure*}[t]
  \centering
  \includegraphics[width=0.75\textwidth]{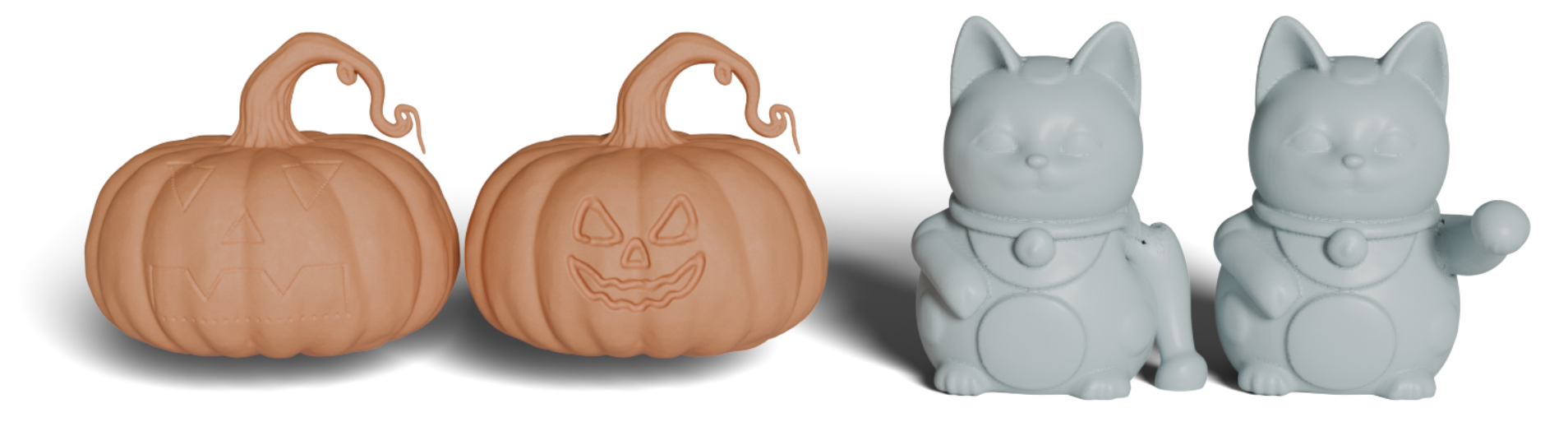}
  \caption{Qualitative comparison with a script-centric approach.
  For each pair of samples, the result generated by Blender MCP Server with Claude Sonnet 4.6 is shown on the left, while the result of our visual-centric method is shown on the right. Our approach facilitates localized edits while preserving the overall structure of the original asset.}
  \label{fig:compare_with_script}
  \Description{}
\end{figure*}

We compare against Blender MCP Server~\cite{blender_mcp_server_2026}, an official Blender Lab MCP connector that exposes Blender's Python API to LLMs.
In our experiments, Blender MCP is connected to Claude Desktop via the Blender Connector, with Claude Sonnet 4.6~\cite{anthropic_sonnet46_2026} as the LLM backend.
We omit LL3M~\cite{lu2025ll3m} from our comparisons as the system is currently inaccessible to the public.
As shown in \cref{fig:compare_with_script}, our visual-centric formulation is particularly effective for non-parametric assets that require localized changes, immediate visual feedback, and minimal disturbance to regions outside the target edit.
We present this comparison through representative object-scale case studies, since task definitions and output formats are not yet standardized across these two paradigms.
\looseness=-1

\paragraph{Comparison with Generative Approaches}
Recent 3D generative models~\cite{Zhang2024, li2024craftsman, yang2025omnipart, Hunyuan3d2025, xiang2025trellis2} have achieved impressive results in synthesizing novel 3D assets.
Our setting differs in its objective: the edited mesh should remain identical to the input except for the localized change specified by the instruction.
To probe this distinction, we construct a generative baseline using a ``Render, Edit, Reconstruct'' pipeline: render the input mesh into an image, edit the image with a text-guided 2D model, and reconstruct a 3D shape from the edited view using a large 3D generative model.
We use Nano Banana Pro~\cite{2023arXiv231211805G} for 2D editing and Hunyuan 2.0~\cite{hunyuan3d22025tencent} for 3D generation.
\looseness=-1

\begin{figure*}[t]
  \centering
  \includegraphics[width=\textwidth,height=0.70\textheight,keepaspectratio]{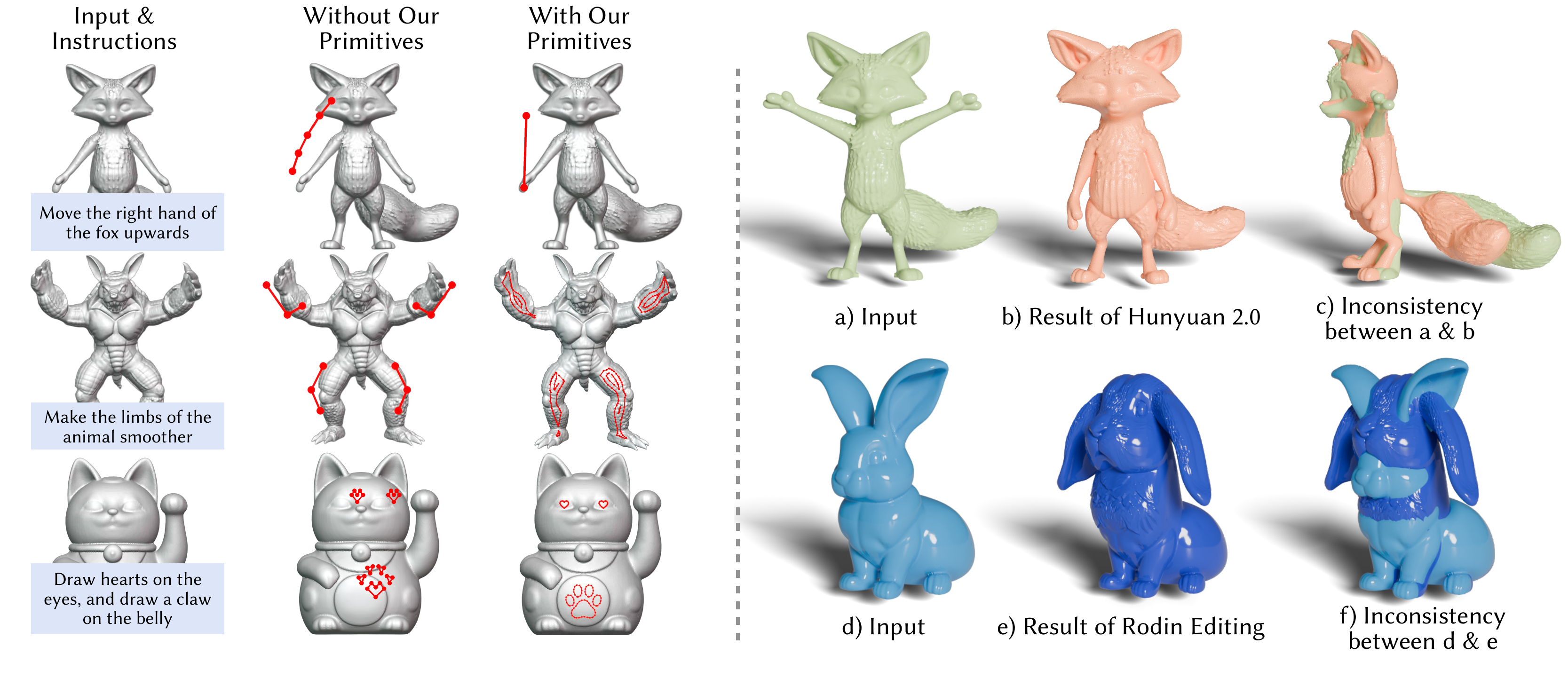}
  \caption{
  \textbf{Left: Ablation study on the effect of our Primitive Abstraction}. Compared to a baseline where the LLM directly outputs raw mouse coordinates, our primitive-based approach effectively grounds semantic instructions into precise, geometrically consistent GUI actions.
  \textbf{Right: Results of generative methods}. We test Hunyuan 2.0 and Rodin on our editing task. For Rodin, we use its editing mode.
    Given the instruction ``Let the fox put its hand down'' and ``Put down the ears of the rabbit'', the generated result exhibits significant geometric deviations from the original input.
  The overlay view (c) and (f) visualizes the structural misalignment between the input and output.}
  \label{fig:ablation_on_primitive_and_compare_with_gen_approach}
\end{figure*}

As shown in the left part of \cref{fig:ablation_on_primitive_and_compare_with_gen_approach}, this pipeline can produce a plausible object, but it often drifts from the original geometry.
The bottleneck is the detour through a single image: projection discards hidden geometry, topology, and fine surface detail, while reconstruction must hallucinate the missing information.
The result is therefore a regenerated mesh conditioned on an edited rendering, rather than an in-place edit of the original asset.
This causes visible changes in regions that should remain untouched.

This comparison highlights the complementary roles of the two paradigms.
Generative models are well suited to creating new shapes or large semantic redesigns, whereas our visual-centric agent directly manipulates the current mesh state through GUI feedback and is therefore better aligned with localized, preservation-critical edits.
Attention-injection techniques~\cite{shen2025qk, hertz2022prompt} may reduce appearance drift in image editing, but they do not remove the geometric information loss introduced once the editing process leaves the native 3D mesh.
\looseness=-1

\paragraph{Ablation: Without Primitive Abstraction}
One of the central design choices of our system is to restrict low-level execution to three primitive mouse actions: \emph{Smear}, \emph{Drag}, and \emph{Draw}.
This abstraction makes the action space tractable for a language-driven agent operating through a GUI.
Without such primitives, the Action Agent would need to directly generate long, high-dimensional free-form mouse trajectories together with brush configurations, localization decisions, and execution timing.

In our formulation, the primitives decouple \emph{what} to do from \emph{how} to trace the stroke, letting the planner express compact intentions while deterministic geometric procedures realize trajectories.
We view these three primitives as a key inductive bias that stabilizes execution and makes our visual-centric pipeline feasible.
To validate this design, we compare our method against a baseline where the language model directly outputs the dense sequence of mouse coordinates needed for the geometric operation.
As shown in \cref{fig:ablation_on_primitive_and_compare_with_gen_approach}, removing the primitives sharply degrades performance: raw-coordinate outputs are often jagged and misaligned with the underlying 3D structure.
\looseness=-1

\begin{figure}[t]
  \centering
  \includegraphics[width=0.9\linewidth]{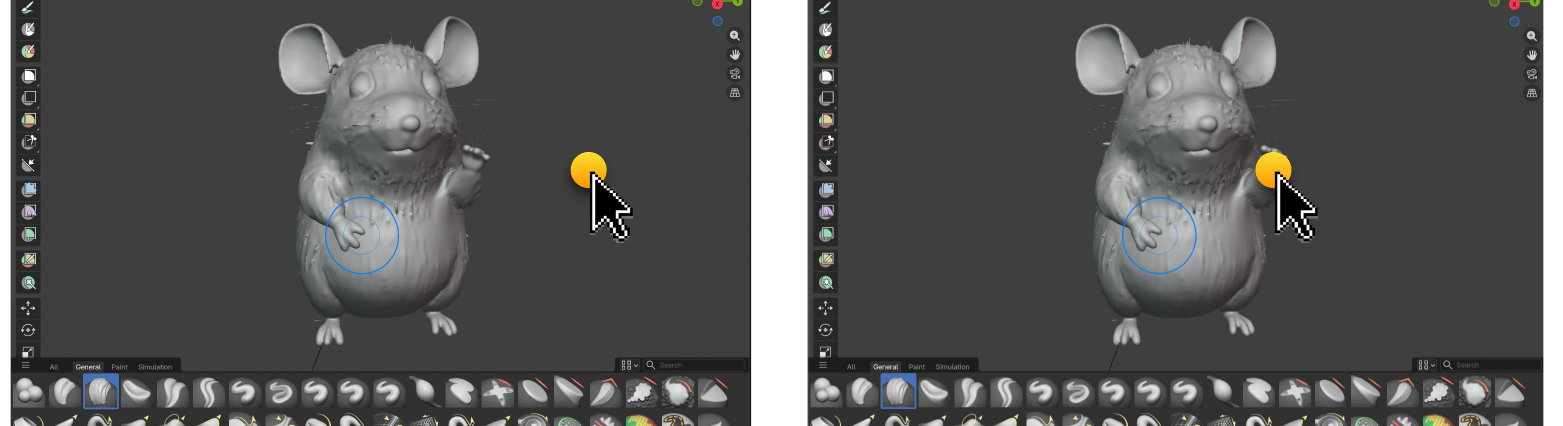}
  \caption{Ablation of QuadLoc. Directly prompting the VLM for coordinates yields inaccurate localization (left), while QuadLoc localizes the target more precisely (right).}
  \label{fig:ablation_quadloc}
  \Description{}
\end{figure}

\paragraph{Ablation: QuadLoc}
We find that our QuadLoc strategy significantly outperforms naive VLM querying. As illustrated in \cref{fig:ablation_quadloc}, directly prompting the VLM to output the coordinates of a specific part (e.g., the mouse's paw) results in inaccurate localization (left). In contrast, with QuadLoc, our method achieves precise localization of the target point (right).

\section{Conclusion} \label{sec:conclusion}

We present a training-free multi-agent framework for 3D mesh editing that emulates the visual-centric workflow of human artists. Unlike script-centric methods, our system interacts directly with the Blender canvas, leveraging vision-language models to interpret user intent and assess progress via visual feedback. We bridge natural language and low-level actions by abstracting sculpting operations into three primitive mouse trajectories: \textit{Smear}, \textit{Drag}, and \textit{Draw}. Our experiments demonstrate promising, intent-aligned localized edits that preserve the original mesh identity.
Rather than claiming universal superiority over script-centric or generative approaches, it represents an alternative paradigm with distinct strengths and trade-offs.
It highlights direct GUI-based interaction as a viable and distinct direction for editing existing 3D assets, particularly when precise in-place editing and region preservation are paramount.

\begin{wrapfigure}[9]{r}{0.5\columnwidth}
\centering
\vspace{-4mm}
   \includegraphics[width=0.95\linewidth]{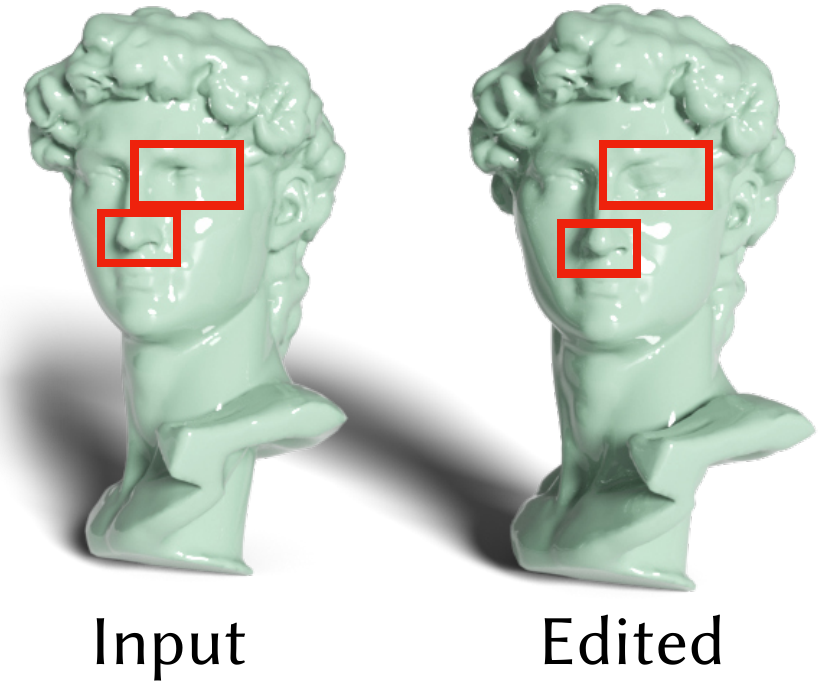}
        \vspace{-4mm}
\end{wrapfigure}

\paragraph{Limitations and Future Work}
Our approach has several limitations that highlight exciting avenues for future research.
First, consistent with prior agent-based frameworks~\cite{lu2025ll3m,yamada2025l3go}, our system heavily relies on underlying foundation models, resulting in latencies of up to 8 minutes per edit. Exploring 3D-native multimodal foundation models could effectively mitigate this issue.
Second, the Planner Agent struggles to translate highly abstract instructions into concrete, multi-step actions. For instance, an ambiguous prompt like ``make David more handsome'' confuses the agent, leading to unpredictable deformations limited to basic \textit{Clay Strip} brush operations, as illustrated in the figure.
Third, despite our multi-view strategies, severe self-occlusions and complex internal geometries challenge our 2D screenshot-based perception. Equipping the agent with autonomous, free-viewpoint camera manipulation will substantially broaden its operational scope.
Fourth, current mouse trajectory primitives are restricted to localized surface deformations. Incorporating \textit{topological primitives} would enable fundamental structural edits, such as boolean operations, drilling holes, or merging objects.
Finally, to address geometric artifacts imperceptible via 2D renderings (e.g., non-manifold edges or self-intersections), future systems could integrate a geometric analysis module into the Reflection Agent, thereby strengthening the alignment between visual plausibility and geometric validity.

\looseness=-1

\bibliographystyle{ACM-Reference-Format}
\bibliography{src/ref/reference}

@String { CAD      = {Computer Aided Design} }

@String { CVPR     = {CVPR} }

@String { ECCV     = {ECCV} }

@String { ICCV     = {ICCV} }

@String { ICLR     = {ICLR} }

@String { ICML     = {ICML} }

@String { NeurIPS  = {NeurIPS} }

@String { SA       = {ACM Trans. Graph. (SIGGRAPH Asia)} }

@String { SGPOLD   = {Symp. Geom. Proc.} }

@String { SIG      = {ACM Trans. Graph. (SIGGRAPH)} }

@String { threeDV  = {3DV} }

@article{Ovsjanikov2012,
  title={Functional maps: a flexible representation of maps between shapes},
  author={Ovsjanikov, Maks and Ben-Chen, Mirela and Solomon, Justin and Butscher, Adrian and Guibas, Leonidas},
  journal=SIG,
  volume={31},
  number={4},
  year={2012},
}

@inproceedings{gao20253d,
  title={3D mesh editing using masked lrms},
  author={Gao, Will and Wang, Dilin and Fan, Yuchen and Bozic, Aljaz and Stuyck, Tuur and Li, Zhengqin and Dong, Zhao and Ranjan, Rakesh and Sarafianos, Nikolaos},
  booktitle=ICCV,
  year={2025}
}

@article{du2024blenderllm,
  title={BlenderLLM: Training Large Language Models for Computer-Aided Design with Self-improvement},
  author={Du, Yuhao and Chen, Shunian and Zan, Wenbo and Li, Peizhao and Wang, Mingxuan and Song, Dingjie and Li, Bo and Hu, Yan and Wang, Benyou},
  journal={arXiv preprint arXiv:2412.14203},
  year={2024}
}

@article{yuan20243d,
  title={{3D}-premise: Can large language models generate {3D} shapes with sharp features and parametric control?},
  author={Yuan, Zeqing and Lan, Haoxuan and Zou, Qiang and Zhao, Junbo},
  journal={arXiv preprint arXiv:2401.06437},
  year={2024}
}

@article{lu2025ll3m,
  title={LL3M: Large language 3D modelers},
  author={Lu, Sining and Chen, Guan and Dinh, Nam Anh and Lang, Itai and Holtzman, Ari and Hanocka, Rana},
  journal={arXiv preprint arXiv:2508.08228},
  year={2025}
}

@inproceedings{hu2024scenecraft,
  title={Scenecraft: An LLM agent for synthesizing {3D} scenes as blender code},
  author={Hu, Ziniu and Iscen, Ahmet and Jain, Aashi and Kipf, Thomas and Yue, Yisong and Ross, David A and Schmid, Cordelia and Fathi, Alireza},
  booktitle=ICML,
  year={2024}
}

@inproceedings{raistrick2024infinigen,
  title={Infinigen indoors: Photorealistic indoor scenes using procedural generation},
  author={Raistrick, Alexander and Mei, Lingjie and Kayan, Karhan and Yan, David and Zuo, Yiming and Han, Beining and Wen, Hongyu and Parakh, Meenal and Alexandropoulos, Stamatis and Lipson, Lahav and others},
  booktitle=CVPR,
  year={2024}
}

@inproceedings{raistrick2023infinite,
  title={Infinite photorealistic worlds using procedural generation},
  author={Raistrick, Alexander and Lipson, Lahav and Ma, Zeyu and Mei, Lingjie and Wang, Mingzhe and Zuo, Yiming and Kayan, Karhan and Wen, Hongyu and Han, Beining and Wang, Yihan and others},
  booktitle=CVPR,
  year={2023}
}

@article{kania2020ucsg,
  title={UCSG-NET-unsupervised discovering of constructive solid geometry tree},
  author={Kania, Kacper and Zieba, Maciej and Kajdanowicz, Tomasz},
  journal=NeurIPS,
  volume={33},
  year={2020}
}

@article{du2018inversecsg,
  title={Inversecsg: Automatic conversion of 3d models to csg trees},
  author={Du, Tao and Inala, Jeevana Priya and Pu, Yewen and Spielberg, Andrew and Schulz, Adriana and Rus, Daniela and Solar-Lezama, Armando and Matusik, Wojciech},
  journal=SA,
  volume={37},
  number={6},
  year={2018},
}

@inproceedings{wu2021deepcad,
  title={Deepcad: A deep generative network for computer-aided design models},
  author={Wu, Rundi and Xiao, Chang and Zheng, Changxi},
  booktitle=CVPR,
  year={2021}
}

@article{chen2025img2cad,
  title={Img2cad: Conditioned 3-d cad model generation from single image with structured visual geometry},
  author={Chen, Tianrun and Yu, Chunan and Hu, Yuanqi and Li, Jing and Xu, Tao and Cao, Runlong and Zhu, Lanyun and Zang, Ying and Zhang, Yong and Li, Zejian and others},
  journal={IEEE Transactions on Industrial Informatics},
  year={2025},
}

@article{jones2020shapeassembly,
  title={Shapeassembly: Learning to generate programs for 3d shape structure synthesis},
  author={Jones, R Kenny and Barton, Theresa and Xu, Xianghao and Wang, Kai and Jiang, Ellen and Guerrero, Paul and Mitra, Niloy J and Ritchie, Daniel},
  journal=SA,
  volume={39},
  number={6},
  year={2020},
}

@inproceedings{yao2022react,
  title={React: Synergizing reasoning and acting in language models},
  author={Yao, Shunyu and Zhao, Jeffrey and Yu, Dian and Du, Nan and Shafran, Izhak and Narasimhan, Karthik R and Cao, Yuan},
  booktitle=ICLR,
  year={2022}
}

@article{alrashedy2024generating,
  title={Generating {CAD} code with vision-language models for 3d designs},
  author={Alrashedy, Kamel and Tambwekar, Pradyumna and Zaidi, Zulfiqar and Langwasser, Megan and Xu, Wei and Gombolay, Matthew},
  journal={arXiv preprint arXiv:2410.05340},
  year={2024}
}

@article{Hunyuan3d2025,
  title   = {Hunyuan3D 2.1: From Images to High-Fidelity 3D Assets with Production-Ready PBR Material},
  author  = {Tencent Hunyuan3D Team},
  year    = {2025},
  journal = {arXiv preprint arXiv:2506.15442}
}

@article{belle2025agents,
  title={Agents of Change: Self-Evolving LLM Agents for Strategic Planning},
  author={Belle, Nikolas and Barnes, Dakota and Amayuelas, Alfonso and Bercovich, Ivan and Wang, Xin Eric and Wang, William},
  journal={arXiv preprint arXiv:2506.04651},
  year={2025}
}

@inproceedings{Hoppe1999,
  title={New quadric metric for simplifying meshes with appearance attributes},
  author={Hoppe, Hugues},
  booktitle={Visualization},
  year={1999},
}

@InProceedings{li2022n,
  title = 	 {{BLIP}: Bootstrapping Language-Image Pre-training for Unified Vision-Language Understanding and Generation},
  author =       {Li, Junnan and Li, Dongxu and Xiong, Caiming and Hoi, Steven},
  booktitle = 	 ICML,
  year = 	 {2022},
}

@article{Krishna2023,
      title={ConceptFusion: Open-set Multimodal 3D Mapping},
      author={Krishna Murthy Jatavallabhula and Alihusein Kuwajerwala and Qiao Gu and Mohd Omama and Tao Chen and Alaa Maalouf and Shuang Li and Ganesh Iyer and Soroush Saryazdi and Nikhil Keetha and Ayush Tewari and Joshua B. Tenenbaum and Celso Miguel de Melo and Madhava Krishna and Liam Paull and Florian Shkurti and Antonio Torralba},
      year={2023},
      Journal={arXiv preprint arXiv:2302.07241},
}

@incollection{Alliez2003,
  title={Anisotropic polygonal remeshing},
  author={Alliez, Pierre and Cohen-Steiner, David and Devillers, Olivier and L{\'e}vy, Bruno and Desbrun, Mathieu},
  booktitle={SIGGRAPH},
  year={2003}
}

@inproceedings{Yan2009,
  title={Isotropic remeshing with fast and exact computation of restricted Voronoi diagram},
  author={Yan, Dong-Ming and L{\'e}vy, Bruno and Liu, Yang and Sun, Feng and Wang, Wenping},
  booktitle={Computer graphics forum},
  volume={28},
  number={5},
  year={2009},
}

@inproceedings{Li2024a,
  title={NASM: Neural Anisotropic Surface Meshing},
  author={Li, Hongbo and Zhu, Haikuan and Zhong, Sikai and Wang, Ningna and Lin, Cheng and Guo, Xiaohu and Xin, Shiqing and Wang, Wenping and Hua, Jing and Zhong, Zichun},
  booktitle=SA,
  year={2024}
}

@inproceedings{mohammad2022,
  title={Clip-mesh: Generating textured meshes from text using pretrained image-text models},
  author={Mohammad Khalid, Nasir and Xie, Tianhao and Belilovsky, Eugene and Popa, Tiberiu},
  booktitle={SIGGRAPH Asia},
  year={2022}
}

@InProceedings{Chen2024s,
    author    = {Chen, Boyuan and Xu, Zhuo and Kirmani, Sean and Ichter, Brain and Sadigh, Dorsa and Guibas, Leonidas and Xia, Fei},
    title     = {SpatialVLM: Endowing Vision-Language Models with Spatial Reasoning Capabilities},
    booktitle = CVPR,
    year      = {2024},
}

@inproceedings{kirillov2023,
  title={Segment anything},
  author={Kirillov, Alexander and Mintun, Eric and Ravi, Nikhila and Mao, Hanzi and Rolland, Chloe and Gustafson, Laura and Xiao, Tete and Whitehead, Spencer and Berg, Alexander C and Lo, Wan-Yen and others},
  booktitle=ICCV,
  year={2023}
}

@article{ravi2024,
  title={Sam 2: Segment anything in images and videos},
  author={Ravi, Nikhila and Gabeur, Valentin and Hu, Yuan-Ting and Hu, Ronghang and Ryali, Chaitanya and Ma, Tengyu and Khedr, Haitham and R{\"a}dle, Roman and Rolland, Chloe and Gustafson, Laura and others},
  journal={arXiv preprint arXiv:2408.00714},
  year={2024}
}

@article{cen2023,
  title={Segment anything in 3D with NeRFs},
  author={Cen, Jiazhong and Zhou, Zanwei and Fang, Jiemin and Shen, Wei and Xie, Lingxi and Jiang, Dongsheng and Zhang, Xiaopeng and Tian, Qi and others},
  journal={Advances in Neural Information Processing Systems},
  volume={36},
  year={2023}
}

@article{yang2023,
  title={Sam3d: Segment anything in 3d scenes},
  author={Yang, Yunhan and Wu, Xiaoyang and He, Tong and Zhao, Hengshuang and Liu, Xihui},
  journal={arXiv preprint arXiv:2306.03908},
  year={2023}
}

@article{liu2025,
      title={Agentic 3D Scene Generation with Spatially Contextualized VLMs},
      author={Xinhang Liu and Yu-Wing Tai and Chi-Keung Tang},
      year={2025},
      journal={arXiv preprint arXiv:2505.20129},
}

@article{fu2024scene,
  title={Scene-llm: Extending language model for 3d visual understanding and reasoning},
  author={Fu, Rao and Liu, Jingyu and Chen, Xilun and Nie, Yixin and Xiong, Wenhan},
  journal={arXiv preprint arXiv:2403.11401},
  year={2024}
}

@inproceedings{wei2024n,
  title={Nto3d: Neural target object 3d reconstruction with segment anything},
  author={Wei, Xiaobao and Zhang, Renrui and Wu, Jiarui and Liu, Jiaming and Lu, Ming and Guo, Yandong and Zhang, Shanghang},
  booktitle=CVPR,
  year={2024}
}

@article{shen2023,
      title={Anything-3D: Towards Single-view Anything Reconstruction in the Wild},
      author={Qiuhong Shen and Xingyi Yang and Xinchao Wang},
      year={2023},
      journal={arXiv preprint arXiv:2310.14261},
}

@article{ling2025,
  title={Scenethesis: A language and vision agentic framework for 3d scene generation},
  author={Ling, Lu and Lin, Chen-Hsuan and Lin, Tsung-Yi and Ding, Yifan and Zeng, Yu and Sheng, Yichen and Ge, Yunhao and Liu, Ming-Yu and Bera, Aniket and Li, Zhaoshuo},
  journal={arXiv preprint arXiv:2505.02836},
  year={2025}
}

@book{Botsch2010,
  author    = {Mario Botsch and Leif Kobbelt and Mark Pauly and Pierre Alliez and Bruno L\'{e}vy},
  title     = {Polygon Mesh Processing},
  publisher = {CRC press},
  year      = {2010}
}

@article{cai2025z,
  title={Z-Image: An Efficient Image Generation Foundation Model with Single-Stream Diffusion Transformer},
  author={Cai, Huanqia and Cao, Sihan and Du, Ruoyi and Gao, Peng and Hoi, Steven and Hou, Zhaohui and Huang, Shijie and Jiang, Dengyang and Jin, Xin and Li, Liangchen and others},
  journal={arXiv preprint arXiv:2511.22699},
  year={2025}
}

@inproceedings{Desbrun1999,
  author    = {Desbrun, Mathieu and Meyer, Mark and Schr\"{o}der, Peter and Barr, Alan H.},
  title     = {Implicit fairing of irregular meshes using diffusion and curvature flow},
  booktitle = {SIGGRAPH},
  year      = {1999}
}

@inproceedings{Garland1997,
  title     = {Surface simplification using quadric error metrics},
  author    = {Garland, Michael and Heckbert, Paul S},
  booktitle = SIG,
  year      = {1997}
}

@article{Gpt4,
  title   = {GPT-4 Technical Report},
  author  = {OpenAI},
  journal = {arXiv preprint arXiv:2303.08774},
  year    = {2023}
}

@article{hertz2022prompt,
  title={Prompt-to-prompt image editing with cross attention control},
  author={Hertz, Amir and Mokady, Ron and Tenenbaum, Jay and Aberman, Kfir and Pritch, Yael and Cohen-Or, Daniel},
  journal={arXiv preprint arXiv:2208.01626},
  year={2022}
}

@inproceedings{Radford2021,
  title     = {Learning transferable visual models from natural language supervision},
  author    = {Radford, Alec and Kim, Jong Wook and Hallacy, Chris and Ramesh, Aditya and Goh, Gabriel and Agarwal, Sandhini and Sastry, Girish and Askell, Amanda and Mishkin, Pamela and Clark, Jack and others},
  booktitle = ICML,
  year      = {2021}
}

@inproceedings{Sharma2018,
  author    = {Gopal Sharma and Rishabh Goyal and Difan Liu and Evangelos Kalogerakis and Subhransu Maji},
  title     = {{CSGNet}: Neural shape parser for constructive solid geometry},
  booktitle = CVPR,
  year      = {2018}
}

@inproceedings{shen2025qk,
  title={QK-Edit: Revisiting Attention-based Injection in MM-DiT for Image and Video Editing},
  author={Shen, Tiancheng and Huang, Zilong and Li, Xiangtai and Lin, Zhijie and Liu, Jiyang and Wang, Yitong and Feng, Jiashi and Yang, Ming-Hsuan and Liew, Jun Hao},
  booktitle=ICCV,
  year={2025}
}

@inproceedings{Sorkine2004,
  author    = {Sorkine, O. and Cohen-Or, D. and Lipman, Y. and Alexa, M. and R\"{o}ssl, C. and Seidel, H.-P.},
  title     = {Laplacian surface editing},
  booktitle = SGPOLD,
  year      = {2004}
}

@inproceedings{Sorkine2007,
  title     = {As-rigid-as-possible surface modeling},
  author    = {Sorkine, Olga and Alexa, Marc},
  booktitle = SGPOLD,
  year      = {2007}
}

@article{Zhang2024,
  title   = {{CLAY}: A Controllable Large-scale Generative Model for Creating High-quality {3D} Assets},
  author  = {Zhang, Longwen and Wang, Ziyu and Zhang, Qixuan and Qiu, Qiwei and Pang, Anqi and Jiang, Haoran and Yang, Wei and Xu, Lan and Yu, Jingyi},
  journal = SIG,
  volume  = {43},
  number  = {4},
  year    = {2024}
}

@inproceedings{Barda2024,
author = {Barda, Amir and Kim, Vladimir and Aigerman, Noam and Bermano, Amit Haim and Groueix, Thibault},
title = {MagicClay: Sculpting Meshes With Generative Neural Fields},
year = {2024},
booktitle = SA,
}

@InProceedings{Barda2025,
    author    = {Barda, Amir and Gadelha, Matheus and Kim, Vladimir G. and Aigerman, Noam and Bermano, Amit H. and Groueix, Thibault},
    title     = {Instant3dit: Multiview Inpainting for Fast Editing of 3D Objects},
    booktitle = CVPR,
    year      = {2025},
}

@software{Ahuja2025blendermcp,
  author = {Siddharth Ahuja and BlenderMCP Contributors},
  title = {{BlenderMCP} - Blender Model Context Protocol Integration},
  year = {2025},
}

@InProceedings{Lv2024,
    author    = {Lv, Jiaxi and Huang, Yi and Yan, Mingfu and Huang, Jiancheng and Liu, Jianzhuang and Liu, Yifan and Wen, Yafei and Chen, Xiaoxin and Chen, Shifeng},
    title     = {GPT4Motion: Scripting Physical Motions in Text-to-Video Generation via Blender-Oriented GPT Planning},
    booktitle = CVPR,
    year      = {2024}
}

@InProceedings{Sun2025,
  author={Sun, Chunyi and Han, Junlin and Deng, Weijian and Wang, Xinlong and Qin, Zishan and Gould, Stephen},
  booktitle=threeDV,
  title={3D-GPT: Procedural 3D Modeling with Large Language Models},
  year={2025}
}

@inproceedings{huang2024,
  title={Blenderalchemy: Editing 3d graphics with vision-language models},
  author={Huang, Ian and Yang, Guandao and Guibas, Leonidas},
  booktitle=ECCV,
  year={2024},
}

@article{ravi2024sam2,
  title={SAM 2: Segment Anything in Images and Videos},
  author={Ravi, Nikhila and Gabeur, Valentin and Hu, Yuan-Ting and Hu, Ronghang and Ryali, Chaitanya and Ma, Tengyu and Khedr, Haitham and R{\"a}dle, Roman and Rolland, Chloe and Gustafson, Laura and Mintun, Eric and Pan, Junting and Alwala, Kalyan Vasudev and Carion, Nicolas and Wu, Chao-Yuan and Girshick, Ross and Doll{\'a}r, Piotr and Feichtenhofer, Christoph},
  journal={arXiv preprint arXiv:2408.00714},
  year={2024}
}

@article{2023arXiv231211805G,
  author = {{Gemini Team}},
  title = "{Gemini: A Family of Highly Capable Multimodal Models}",
  journal = {arXiv e-prints},
  year = 2023,
}

@article{hunyuan3d22025tencent,
    title={Hunyuan3D 2.0: Scaling Diffusion Models for High Resolution Textured 3D Assets Generation},
    author={Tencent Hunyuan3D Team},
    year={2025},
}

@article{li2024craftsman,
  author    = {Weiyu Li and Jiarui Liu and Hongyu Yan and Rui Chen and Yixun  Liang and Xuelin Chen and Ping Tan and Xiaoxiao Long},
  title     = {CraftsMan3D: High-fidelity Mesh Generation with 3D Native  Generation and Interactive Geometry Refiner},
  journal   = {arXiv preprint arXiv:2405.14979},
  year      = {2024},
}

@article{yang2025omnipart,
  title={Omnipart: Part-aware 3d generation with semantic decoupling and structural cohesion},
  author={Yang, Yunhan and Zhou, Yufan and Guo, Yuan-Chen and Zou, Zi-Xin and Huang, Yukun and Liu, Ying-Tian and Xu, Hao and Liang, Ding and Cao, Yan-Pei and Liu, Xihui},
  journal={arXiv preprint arXiv:2507.06165},
  year={2025}
}

@article{xiang2025trellis2,
    title={Native and Compact Structured Latents for 3D Generation},
    author={Xiang, Jianfeng and Chen, Xiaoxue and Xu, Sicheng and Wang, Ruicheng and Lv, Zelong and Deng, Yu and Zhu, Hongyuan and Dong, Yue and Zhao, Hao and Yuan, Nicholas Jing and Yang, Jiaolong},
    journal={Tech report},
    year={2025}
}

@techreport{anthropic2025claudesonnet45,
  title={System Card: Claude Sonnet 4.5},
  author={Anthropic},
  year={2025},
  institution={Anthropic}
}

@techreport{seed2025seed18,
  title={Seed1.8 Model Card: Towards Generalized Real-World Agency},
  author={ByteDance Seed},
  year={2025},
  institution={ByteDance}
}

@article{Glm4.6v,
      title={GLM-4.5V and GLM-4.1V-Thinking: Towards Versatile Multimodal Reasoning with Scalable Reinforcement Learning},
      author={GLM},
      year={2025},
      journal={arXiv preprint arXiv:2507.01006},
}

@techreport{openai2025gpt52,
  title={Update to GPT-5 System Card: GPT-5.2},
  author={OpenAI},
  year={2025},
  institution={OpenAI}
}

@techreport{google2025gemini3pro,
  author = {Google},
  institution = {Google},
  title = {Gemini 3 Pro Model Card},
  year = {2025}
}

@techreport{xai2025grok4,
  author = {xAI},
  institution = {xAI},
  title = {Grok 4 Model Card},
  year = {2025}
}

@techreport{xai2025grok41,
  author = {xAI},
  institution = {xAI},
  title = {Grok 4.1 Model Card},
  year = {2025}
}

@article{Qwen3-VL,
  title={Qwen3-VL Technical Report},
  author={Qwen},
	journal={arXiv preprint arXiv:2511.21631},
  year={2025}
}

@inproceedings{yamada2025l3go,
    title = "{L}3{GO}: Language Agents with Chain-of-3{D}-Thoughts for Generating Unconventional Objects",
    author = "Yamada, Yutaro and Chandu, Khyathi and Lin, Bill Yuchen and Hessel, Jack and Yildirim, Ilker and Choi, Yejin",
    booktitle = "Proceedings of the 2025 Conference of the Nations of the Americas Chapter of the Association for Computational Linguistics: Human Language Technologies (System Demonstrations)",
    year = "2025",
}

@misc{blender_mcp_server_2026,
  author       = {{Blender Foundation}},
  title        = {{Blender MCP Server}},
  year         = {2026},
  howpublished = {\url{https://projects.blender.org/lab/blender_mcp}},
}

@misc{anthropic_sonnet46_2026,
  author       = {{Anthropic}},
  title        = {{Introducing Claude Sonnet 4.6}},
  year         = {2026},
  howpublished = {\url{https://www.anthropic.com/news/claude-sonnet-4-6}},
}

\end{document}